\documentclass[preprint,journal]{vgtc}       

\ifpdf
  \pdfoutput=1\relax                   
  \usepackage{graphicx}                
  \DeclareGraphicsExtensions{.pdf,.png,.jpg,.jpeg} 
\else
  \usepackage{graphicx}                
  \DeclareGraphicsExtensions{.eps}     
\fi%

\graphicspath{{figures/}{pictures/}{images/}{./}} 

\usepackage{microtype}                 
\PassOptionsToPackage{warn}{textcomp}  
\usepackage{textcomp}                  
\usepackage{mathptmx}                  
\usepackage{enumitem}

\usepackage{times}                     
\usepackage{cite}                      
\usepackage{tabu}                      
\usepackage{booktabs}                  
\usepackage{multirow}
\usepackage{amsmath}
\usepackage{amssymb}
\usepackage{xcolor}
\usepackage{hhline}

\onlineid{1494}

\vgtccategory{Research}
\vgtcpapertype{algorithm/technique}

\title{Interactive Visual Pattern Search on Graph Data via Graph Representation Learning}

\author{Huan Song, Zeng Dai*, Panpan Xu*, Liu Ren}
\authorfooter{
\item
 Huan Song and Liu Ren are with Robert Bosch Research and Technology Center, USA. E-mail: huan.song, liu.ren@us.bosch.com.
\item
 Zeng Dai is with ByteDance Inc. E-mail: zeng.dai@bytedance.com.
\item
 Panpan Xu is with Amazon AWS AI. E-mail: xupanpan@amazon.com.
\item[]
* Work done while the authors were with Robert Bosch Research and Technology Center, USA.
}

\newcommand{\del}[1]{}

\newcommand{\systemname}{GraphQ}
\newcommand{\algorithmname}{NeuroAlign}

\abstract{
Graphs are a ubiquitous data structure to model processes and relations in a wide range of domains. 
Examples include control-flow graphs in programs and semantic scene graphs in images. Identifying subgraph patterns in graphs is an important approach to understand their structural properties. We propose a visual analytics system \systemname~to support human-in-the-loop, example-based, subgraph pattern search in a database containing many individual graphs. 
{To support fast, interactive queries, we use graph neural networks (GNNs) to encode a graph as fixed-length latent vector representation, and perform subgraph matching in the latent space. Due to the complexity of the problem, it is still difficult to obtain accurate one-to-one node correspondences in the matching results that are crucial for visualization and interpretation. We, therefore, propose a novel GNN for node-alignment called \algorithmname, to facilitate easy validation and interpretation of the query results.} \systemname~provides a visual query interface with a query editor and a multi-scale visualization of the results, as well as a user feedback mechanism for refining the results with additional constraints. We demonstrate \systemname~through two example usage scenarios: analyzing reusable subroutines in program workflows and semantic scene graph search in images. Quantitative experiments show that \algorithmname~achieves 19\%--29\% improvement in node-alignment accuracy compared to baseline GNN and provides up to 100x speedup compared to combinatorial algorithms. {Our qualitative study with domain experts confirms the effectiveness for both usage scenarios.}\looseness=-1
} 


\teaser{
  \centering
  \includegraphics[width=0.9\linewidth]{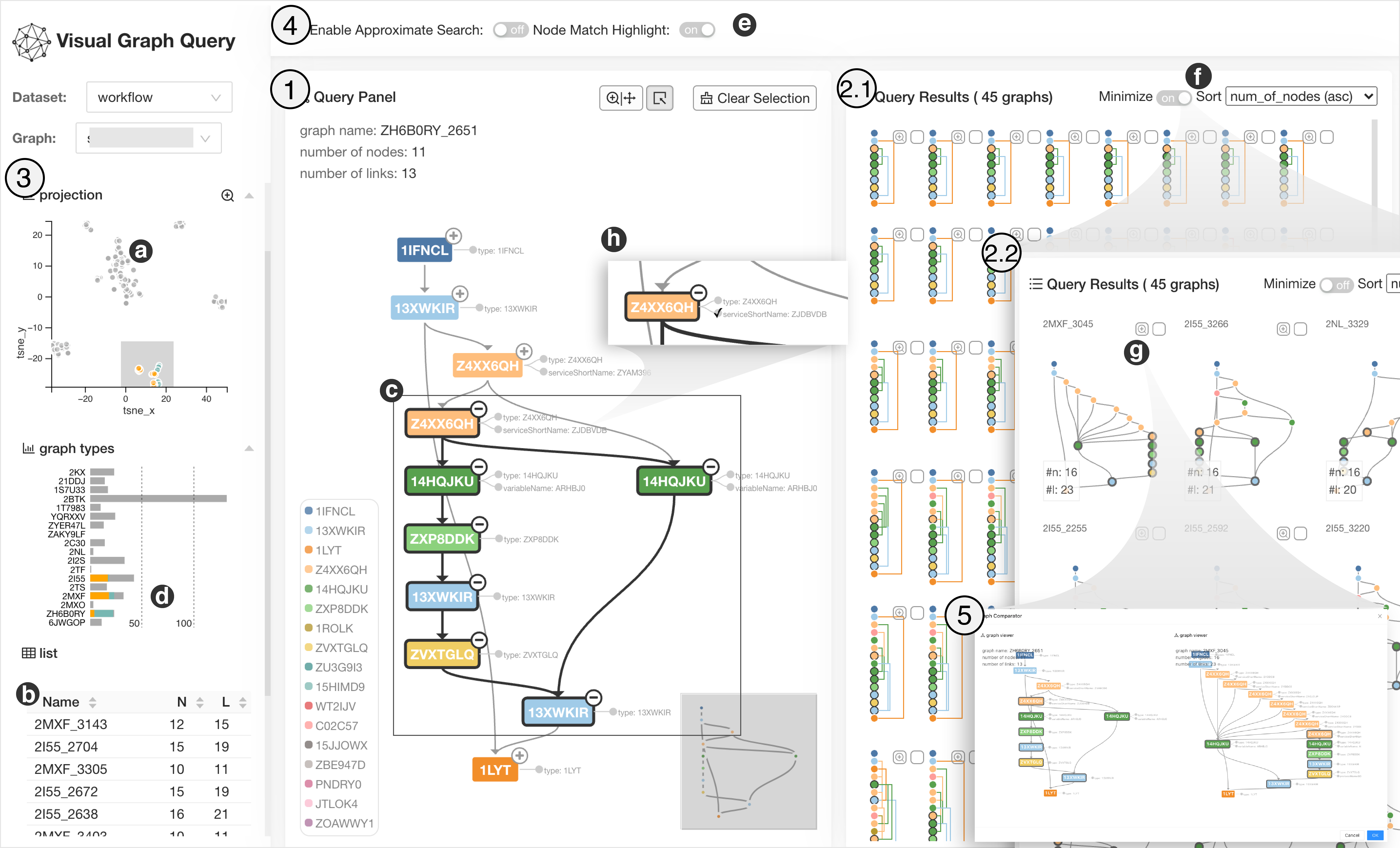}
  \caption{The visualization interface of \systemname~contains: (1) A query editing panel to specify the subgraph patterns and initiate the search. (2.1) (2.2) Query result panels to display the retrieved results. The graph thumbnails can be displayed in overview and detail modes. (3) A statistics and filtering panel that helps users select a graph to construct example-based query, and visualizes the distribution of the query results in the database. (4) A query option control panel to specify whether fuzzy-pattern search is enabled and whether the node-match should be highlighted. (5) A popup window for pairwise comparison between the query pattern and the returned result. The figure shows a case study on program workflow graph pattern search and the details are described in Section~\ref{subsection:workflow_analysis}.}\looseness=-1
  \label{fig:teaser}
}

\vgtcinsertpkg

\ieeedoi{10.1109/TVCG.2021.3114857}

\begin{document}


\firstsection{Introduction}

\maketitle

The graph data structure models a wide range of processes and relations in real-world applications. Examples include business processes\cite{streit2005visualization}, control flow graphs in programs\cite{allen1970controlflow}, social connections \cite{wasserman1994social, perer2006balancing}, knowledge graphs \cite{ji2020survey} and semantic scene graphs in image analysis \cite{neumann2016propagation}. Visually identifying and searching for persistent subgraph patterns is a common and important task in graph analysis. For example, searching for graph motifs such as cliques or stars in a social network reveals the community structures or influencers \cite{dunne2013motif}; searching for similar workflow templates helps streamline or simplify business processes; searching for images with similar scene graphs helps systematic retrieval of training/testing cases to develop models for computer vision tasks.\looseness=-1

In this work, our goal is to support human-in-the-loop, example-based graph pattern search in a graph database, which could contain hundreds to thousands of individual graphs. Supporting interactive, example-based visual graph pattern query is challenging. Previous graph motif/pattern finding algorithms, {e.g. \cite{vehlow2017groupstructure,pienta2016visage,pienta2017vigor}} often impose a strict limit on the size of query pattern and do not scale well as the size of the query pattern and the number or the size of the query targets increases. In fact, subgraph matching is a well-known NP-complete problem \cite{ullmann1976algorithm} and there is no known efficient solution so far.  Furthermore, the complexity of the subgraph matching problem also makes it difficult to obtain accurate one-to-one node correspondence in the matching results. {The node correspondences are crucial to enable visualization-based interpretation and verification of the model's finding.} Besides that, it is quite often that domain knowledge is needed to further refine and adjust the results, which cannot be easily supported in algorithms with heavy computational costs.\looseness=-1

To address those challenges, we propose a novel framework for interactive visual graph pattern search via graph representation learning. Our approach leverages graph neural networks (GNNs) to encode topological as well as node attribute information in a graph as fixed-length vectors. The GNNs are applied to both the query graph and the query targets to obtain their respective vector representations. The graph matching problem is therefore transformed into a high-dimensional vector comparison problem, which greatly reduces the computational complexity. In particular, we leverage two separate GNNs to address 1) the \textbf{\textit{decision problem}} to determine whether a query pattern exists in a graph and 2) the \textbf{\textit{node-alignment problem}} to find the one-to-one node correspondence between the query pattern and the query targets. We leverage NeuroMatch \cite{lou2020neural} for the decision problem. For the node-alignment problem, we propose a novel approach called \algorithmname~that can directly generate cross-graph node-to-node attention scores indicating the node correspondences. In most application scenarios we can precompute and store the vector representations of the query targets for efficient retrieval of the graph matching results. The visualization interface enables easy search and specification of the graph query patterns. Since the query engine could return a large number of matched graphs, we present the results with different levels-of-details that show the matched graphs in space-efficient, thumbnail style representations. They can also be sorted via a variety of criteria. Users can also interactively specify additional constraints to further filter the returned results based on their domain knowledge.\looseness=-1

{We develop the visual analytics system GraphQ based on the proposed framework. GraphQ} goes beyond looking for a predefined set of graph motifs and the users can interactively specify and search for meaningful graph patterns in the respective application domain. The query pattern can include both topological structures and domain-specific node attributes to be matched in the query results. The specified query can be partially matched to enable fuzzy-pattern search.\looseness=-1

We demonstrate \systemname's usefulness with two example usage scenarios in different application domains. In the first usage scenario, we apply the system to analyze a large collection of engineering workflow graphs describing the diagnostics programs in automotive repair shops. The goal is to understand whether there are repetitive patterns in the workflow graphs which eventually serves two purposes -- curate the workflows to reduce repetitive operations and reuse the patterns as templates for future workflow creation.  In the second usage scenario, we apply \systemname~to analyze the semantic scene graphs generated from images, where the nodes are image regions (super-pixels) with semantic labels such as \textit{buildings} and \textit{road}, and the links describe the adjacency relations between regions. Searching for subgraph patterns in such semantic scene graphs can help retrieve similar test cases for model diagnostics in computer vision tasks. The example usage scenarios demonstrate that the framework is generalizable and can be applied to graphs of different nature.\looseness=-1

Furthermore, we conduct quantitative experiments to evaluate the accuracy and the speed of both NeuroMatch and \algorithmname. We show that for the node alignment problem, \algorithmname~can produce 19\%--29\% more accurate results compared to the baseline technique described in NeuroMatch \cite{lou2020neural}. The improvement greatly helps in validating and interpreting the query results in the visualization. We also compared the speed of the algorithm with a baseline combinatorial approach, the result shows that our algorithm gains up to 100$\times$ speed improvement. {The speed improvement is the key that enables a human-in-loop, visual analytics pipeline.}\looseness=-1

To summarize, our contributions include:
\vspace{-0.08in}
\begin{itemize}
    \itemsep-0.3em 
    \item A visual analytics framework for human-in-the-loop, example-based graph pattern search via graph representation learning. {To the best of our knowledge, this is the first deep learning-based approach for interactive graph pattern query.}\looseness=-1
    \item A novel approach (\algorithmname) for pairwise node-alignment based on graph representation learning which provides 10$\times$--100$\times$ speedup compared to baseline combinatorial algorithm \cite{munkres1957algorithms} and 19\%--29\% more accurate results than existing deep learning based approach.\looseness=-1
    \item A prototype implementation of the framework, {GraphQ,} with interactive query specification, query result display with multiple levels-of-detail, and user feedback mechanisms for query refinement. Two example usage scenarios illustrating the general applicability and effectiveness of the proposed {system}.\looseness=-1
\end{itemize}

\section{Related Work}

{In this section, we focus on the most relevant research to our work in the areas of graph visualization, visual graph query, and graph representation learning for subgraph pattern matching.}\looseness=-1

\subsection{Graph Visualization}

Graph visualization is an extensively studied topic \cite{herman2000graph, nobre2019starmultivariategraph} for its application in a wide range of domains. Open source or commercial software for graph visualization (e.g. Gelphi \cite{bastian2009gephi} and Neo4j Bloom\cite{neo4j}) are also available for off-the-shelf use. Researchers in graph visualization typically focus on one or more of the following aspects: develop layout algorithms to efficiently compute readable and aesthetic visualizations {(e.g. \cite{gansner1993technique, bennett2007aesthetics, diaz2002survey,hu2005efficient, jacomy2014forceatlas2, kwon2017would})}, design new visual encoding to display nodes and edges (e.g. \cite{herman2000graph, henry2007nodetrix, van2015reducing}), develop graph simplification or sampling technique to avoid over-plotting and visual clutter (e.g. \cite{dunne2013motif, van2014multivariate}), and design novel user interaction scheme for exploratory analysis {(e.g. \cite{herman2000graph, tominski2006fisheye, pister2020integrating, srinivasan2017graphiti})}. Depending on the nature of the graph data, {they have developed a variety of systems and algorithms for} directed/undirected graphs, multivariate graphs (with node/edge attributes) and dynamic network visualization to support a wide range of graph analytic tasks \cite{lee2006task, pretorius2014tasks}. \looseness=-1

In this work, we focus on supporting interactive, example-based visual query of graph patterns in a database and visualizing the results. This is a generic framework that can be applied to both directed or undirected graph and graphs with node/edge attributes, as demonstrated in the example usage scenarios. We utilize existing graph layout techniques for a detailed view of directed graphs \cite{gansner1993technique} and design a compact visualization for summarizing graph structure to provide an overview of the query results. \looseness=-1

\subsection{{Visual Graph Query}}

Graph patterns/motifs are frequently used to simplify the display of graphs and reduce visual clutter. Motif Simplification \cite{dunne2013motif} was developed to identify graph motifs including \textit{clique}, \textit{fan}, and \textit{d-connectors} based on topological information and visualized them as glyphs in the node-link display for more efficient usage of the screen space. More generally, cluster patterns, esp. ``near-clique'' structures are the most studied and visualized in the literature and various methods have been developed to compute and visualize them \cite{vehlow2017groupstructure}. However, most of the patterns/ motifs here are predefined and can not be easily modified by users.

Graphite\cite{chau2008graphite}, Vogue \cite{bhowmick2013vogue}, and Visage \cite{pienta2016visage} support interactive, user-specified queries on graph data and Vigor \cite{pienta2017vigor} focuses on visualization of the querying results. In these systems, users can interactively specify node attributes as well as topological constraints in the form of a query graph and the system searches for matching subgraphs. However, the complexity of the query is usually limited, which  reduces the expressive power of the specified patterns.\looseness=-1

Our approach is also inspired by a number of existing visual query system on time series data, where the user can interactively specify the patterns they are searching for, by either drawing the pattern directly on a canvas or selecting the pattern from a data sample \cite{wattenberg2001sketch, hochheiser2003interactive, hochheiser2004dynamic, buono2005interactive, lekschas2020peax}. Supporting user-specified patterns gives the user great flexibility and power to perform exploratory analysis in various application domains. However, querying arbitrary patterns on a graph structure brings unique challenges in terms of the computation speed needed to support an interactive user experience, which we address with a graph representation learning-based approach.\looseness=-1

\subsection{Graph Representation Learning for Subgraph Pattern Matching}

Graph neural networks (GNNs) have emerged as a generic approach for graph representation learning, which can support a variety of graph analytics tasks including link prediction, node classification, and community structure identification \cite{kipf2016semi,hamilton2017inductive,velivckovic2017graph,xu2018powerful,shanthamallu2019gramme}. The recent development on GNN library further increases the popularity among researchers \cite{torch_geometric}. The success of GNN on diverse graph tasks also motivated researchers to address the comparison problem between different graphs, such as graph matching \cite{li2019graph} and graph similarity learning \cite{al2019ddgk}. A comprehensive survey on this topic is provided in \cite{ma2019deep}. Recently, GNNs have been shown to improve the performance on the challenging subgraph-isomorphism problems, including subgraph matching \cite{lou2020neural}, subgraph isomorphism counting \cite{liu2020neural}, maximum common subgraph detection \cite{bai2019neural}, and {graph alignment \cite{fey2020deep}. Powered by flexible representation learning, these approaches addressed issues of heuristic-based solutions \cite{heimann2018regal,sun2012efficient} in terms of accuracy and query scalability. Our objective is to utilize GNNs to facilitate fast user-interaction with graph queries, where the embeddings of the existing graphs can be pre-computed and stored to enable efficient retrieval during the inference stage. Compared to \cite{bai2019neural,fey2020deep}, our approach resolves subgraph isomorphism from the learned embedding space alone, without expensive iterative search \cite{bai2019neural} or embedding refinement aided by the additional network \cite{fey2020deep}.} Our proposed {framework} utilizes NeuroMatch \cite{lou2020neural} as a core component to efficiently query matching graphs but involves a novel component \algorithmname\ to resolve the issue of NeuroMatch on obtaining accurate node alignment. The capability to identify matching nodes is critical for intuitive user interaction with complex topologies.\looseness=-1

There are relatively fewer works in the visual analytics domain utilizing graph representation learning. In \cite{fujiwara2020visualconstrastive}, a contrastive learning approach is developed to visualize graph uniqueness and explain learned features. Graph representation learning-based algorithms have also been developed for graph layout/drawing \cite{wang2019deepdrawing, kwon2019deep}, evaluating graph visualization aesthetics  \cite{haleem2019evaluating}, and sample large graphs for visualization \cite{zhou2020context}. Our framework addresses the important problem of subgraph matching and facilitates intuitive interaction. To the best of our knowledge, this is the first approach based on representation learning for interactive visual graph queries. \looseness=-1

\section{Algorithm}

\begin{figure}[t]
	\centering
    \includegraphics[width=0.65\linewidth]{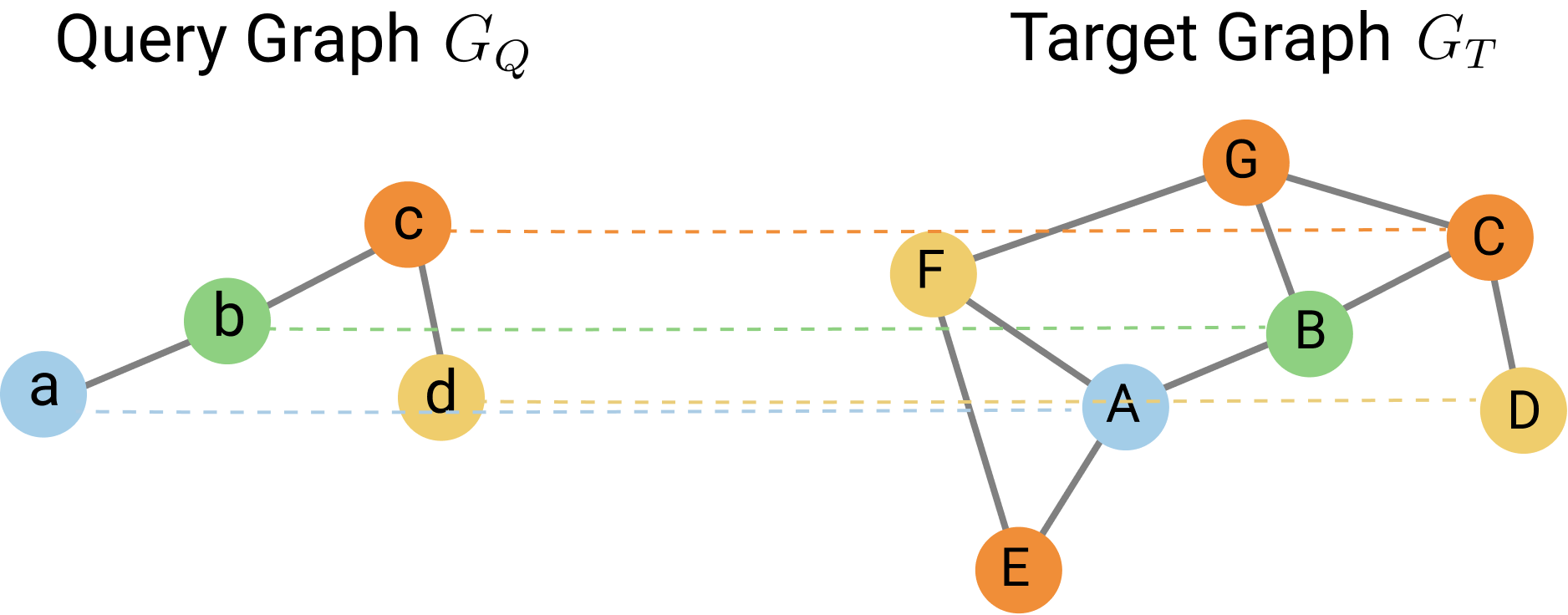}
    \vspace{-0.08in}
    \caption{Visual illustration of the subgraph matching problem. We color-encode the node categorical features of both graphs. The example query graph is subgraph-isomorphic to the target graph with the correct node alignment indicated by dashed lines.\looseness=-1}
    \vspace{-0.10in}
    \label{fig:subgraph}
    \vspace{-0.15in}
    
\end{figure}

{In this section, we first define the subgraph matching problem and describe our overall framework to resolve it. We then describe NeuroMatch and NeuroAlign, the two GNNs as the core components of the framework. Finally, we introduce an improved inference method and a simple extension to support approximate query matching.}\looseness=-1

\subsection{{Problem Definition}}

We first formally define the subgraph matching problems. We denote $G=(V,E)$ as an undirected, connected graph with vertex set $V$ and edge set $E$, $X$ as the features associated with $V$ (e.g.\ categorical attributes). Given a query graph $G_Q$ and a target graph $G_T$, we consider the \textbf{\textit{decision problem}} which determines whether there exists a subgraph $H_T\subseteq G_T$, such that $G_Q$ is isomorphic to $H_T$. When $H_T$ exists, i.e.\ $G_Q$ is subgraph-isomorphic to $G_T$, we further consider the \textbf{\textit{node alignment problem}} which looks for an injective mapping function $f:V_{Q}\rightarrow V_T$, such that $\{f(v),f(u)\}\in E_T$ if $\{v,u\}\in E_{Q}$. When the node features $X$ exist, the matching requires equivalence of the feature too. Note that this defines \textit{edge-induced} subgraph isomorphism, which is our focus in the paper. However, the system is general to apply on \textit{node-induced} subgraph isomorphism \cite{bachl1999isomorphic} too.\looseness=-1

An illustrative example is shown in \autoref{fig:subgraph}, where the colors encode node categorical feature and letters are the node names. The example query graph $G_Q$ is a subgraph of $G_T$ with the correct node alignment of $f(a)=A,f(b)=B,f(c)=C,f(d)=D$. In this paper, we consider the practical case of a large database of target graphs, where the task is to solve the above decision problem and node-alignment problem for each of the target graphs.\looseness=-1

\subsection{{Overall Framework}}

\begin{figure}[t]
	\centering
 	\vspace{-0.15in}
    \includegraphics[width=0.97\linewidth]{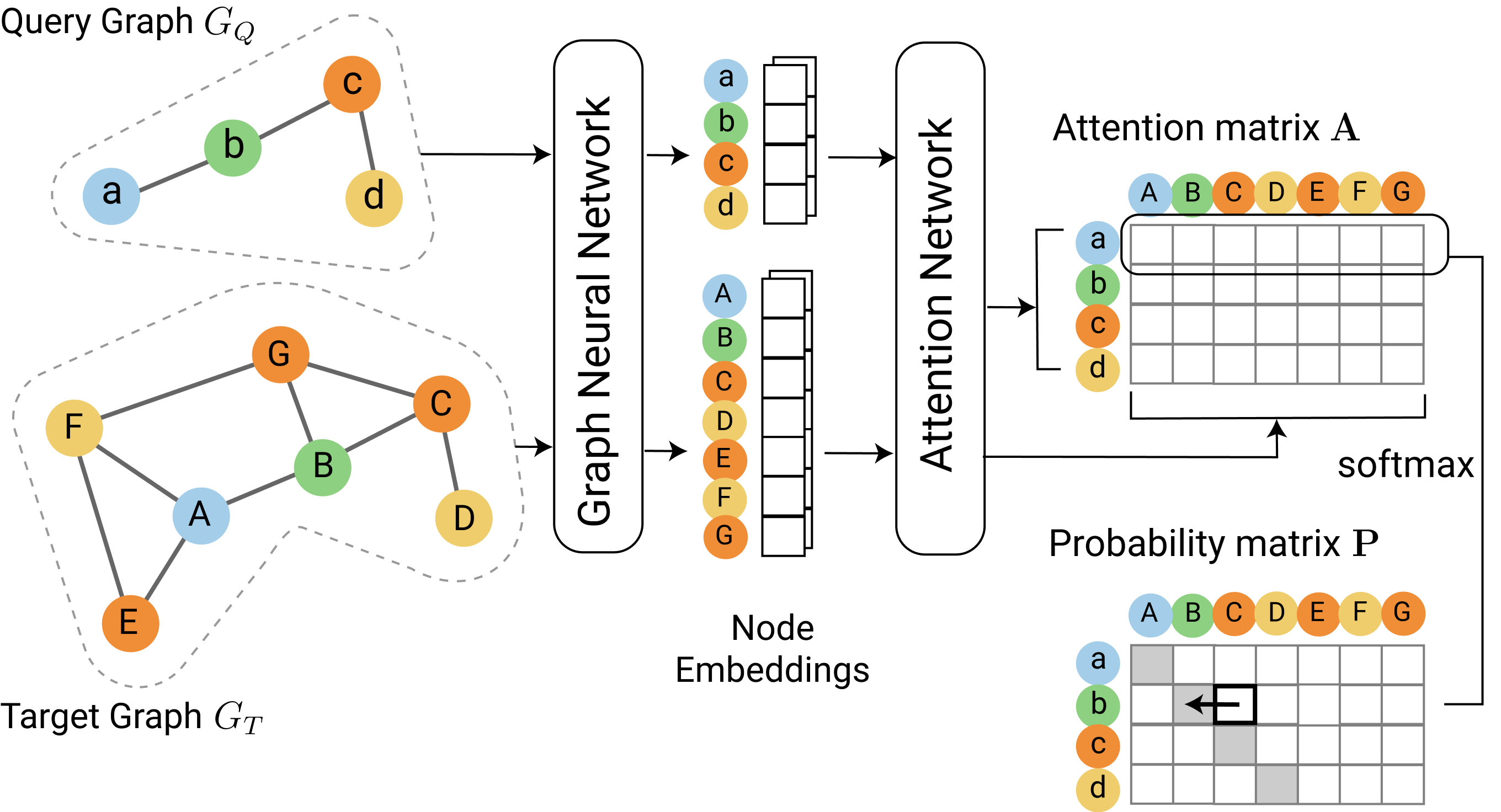}
    \vspace{-0.15in}
    \caption{NeuroAlign algorithm obtains accurate node-to-node correspondence. It extracts the embeddings of each node in the query graph and the target graph by directly feeding them through GNN. It then uses an attention network to compare every pair of node embeddings between the query and target graphs. For the convenience of computation, these pair-wise comparison results are formed as a matrix. The rows correspond to query nodes and columns correspond to target nodes. The matrix is then transformed into a probability matrix through softmax on each row. A greedy assignment algorithm resolves potential conflicts (black outlined block) during inference {(Section \ref{sec:assignment})}.\looseness=-1}
    \label{fig:neuroalign}
    \vspace{-0.10in}
\end{figure}

{Our proposed framework consists of two core components:} NeuroMatch (\autoref{fig:neuromatch}) and NeuroAlign (\autoref{fig:neuroalign}), which focus on solving the subgraph decision and node alignment problems respectively. Given a graph database and user-created query graph, we utilize the state-of-the-art NeuroMatch method \cite{lou2020neural} to efficiently retrieve matching target graphs which contain the query graph. NeuroMatch decomposes the graphs into small neighborhoods to make fast decision locally and then aggregates the results. {After a matching target graph is found, the node alignment between the two graphs can still be ambiguous and misleading based on what we observe in the experimental results. This is due to the fact that the learning process of NeuroMatch relies entirely on small neighborhoods within the graphs. As a result, each query node could end up matched to multiple target nodes where many of them are actually false positives. To tackle these issues, we propose a novel model \algorithmname, which directly predicts node alignment from query and target graphs, without segmenting them into small neighborhoods. It computes node-to-node attention based on graph node embeddings to obtain the alignment results.} Finally, the matching target graphs and corresponding matching nodes are returned to the user for exploration and analysis.  \looseness=-1

NeuroMatch and NeuroAlign both employ GraphSAGE \cite{hamilton2017inductive} as the backbone GNN for representation learning. For simplicity, we consider GraphSAGE as a general function that performs representation learning, where the input is a given graph and the output is a set of embeddings for every node in the graph. Optionally, a pooling layer can be added on top of the node embeddings to obtain a single embedding of the input graph. A more detailed description can be found in the appendix. We use $h_v$ to denote the learned representation of node $v$ at the final output layer, which will be used by NeuroMatch and NeuroAlign as described in the following sections. \looseness=-1

\subsection{{Subgraph Decision via NeuroMatch}}
\label{sec:neuromatch}

\begin{figure}[t]
	\centering
    \vspace{-0.15in}
    \includegraphics[width=0.8\linewidth]{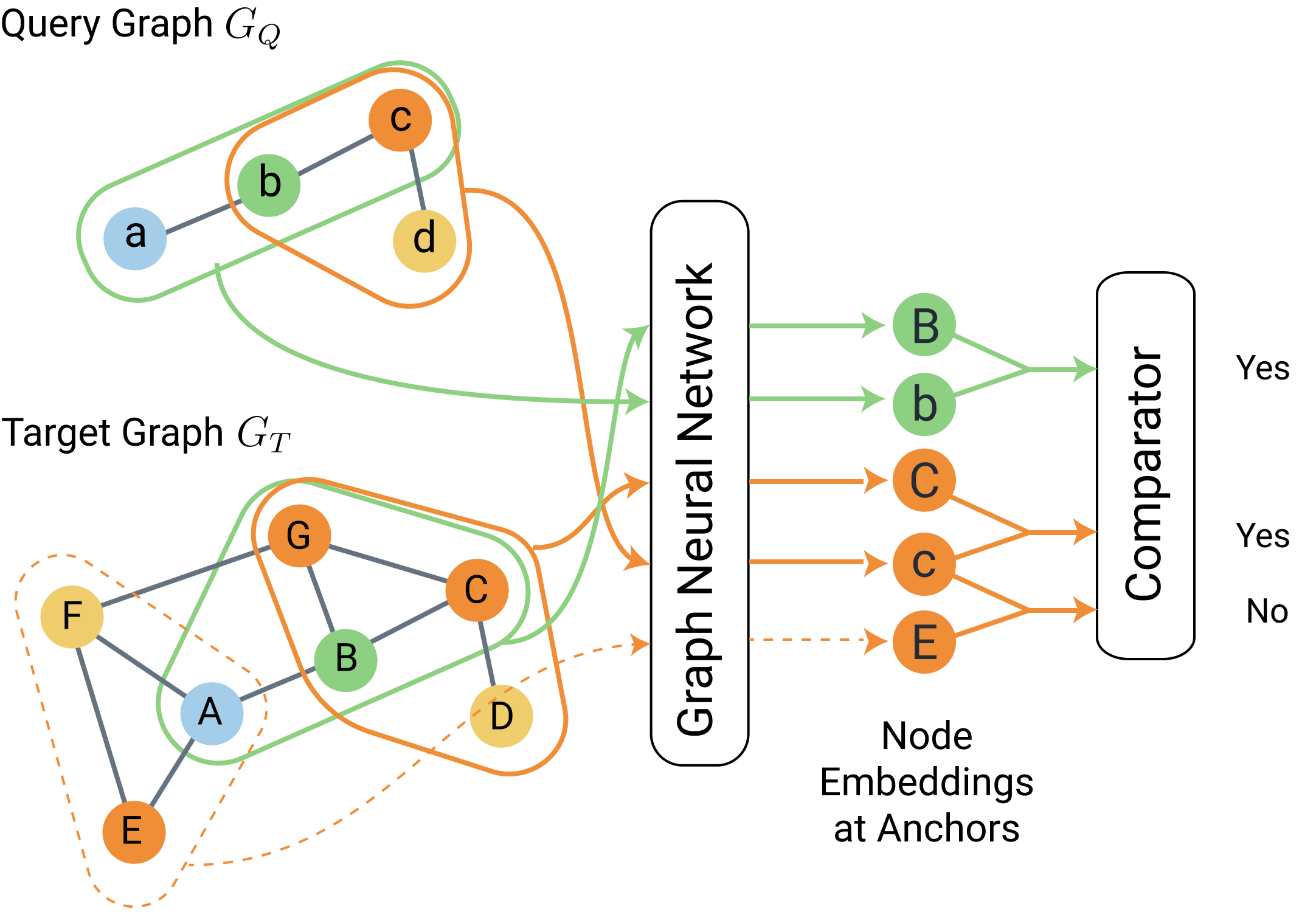}
    \vspace{-0.15in}
    \caption{NeuroMatch determines whether $G_Q$ is a subgraph of $G_T$ by looking for local matches first and then aggregate the results. In this figure, we highlight the $1$-hop local neighborhoods at anchor nodes $b,c$ in the query graph as an example (in green and orange outlines). The NeuroMatch algorithm compares these $1$-hop neighborhoods with those in the target graph. It finds that the $1$-hop neighborhood graph of $b$ is a subgraph of the $1$-hop neighborhood of $B$ (highlighted in green) and the neighborhood of $c$ is a subgraph of the neighborhood of $C$ (highlighted in orange). Since for each query node ($a$, $b$, $c$, $d$), we can find a matching $1$-hop neighborhood graph in the target graph ($A$, $B$, $C$, $D$), the algorithm concludes that indeed $G_Q$ is a subgraph of $G_T$. \looseness=-1}

    \label{fig:neuromatch}
    \vspace{-0.10in}
\end{figure}






{Conducting subgraph matching in the embedding space can facilitate efficient retrieval. However, considering the scale of the database and the large size of certain graphs, it is challenging to build the predictive model to encode the subgraph relationships. NeuroMatch resolves this issue by decomposing the given query and target graphs into many small regions and learns the subgraph relationship in these small regions first.} In particular, for each node $q$ in the query graph, it extracts a small $k$-hop neighborhood graph $g_q$. For each node $t$ in the target graph, it also extracts their $k$-hop neighborhood $g_t$. Then the problem of determining whether $G_Q\subseteq G_T$ transforms into many local subgraph matching decisions about whether $g_q\subseteq g_t$. To find potential local matches, NeuroMatch compares all pairs of nodes between the query and target graphs. Finally, the ensemble decision can be made by checking whether every query neighborhood can find a matching target neighborhood. Figure \ref{fig:neuromatch} shows a simple example to illustrate the main idea of NeuroMatch. In order to determine the local subgraph relationship, i.e.\ whether the $k$-hop neighborhood graph $g_q$ is a subgraph of $g_t$, the algorithm feeds $g_q$ and $g_t$ into GNN with the pooling layer to extract the respective anchor node embedding at $q$ and $t$. A comparator function then takes each pair of these embeddings and predicts the subgraph relationship, as shown in \autoref{fig:neuromatch}. We describe the method in the appendix and refer readers to the NeuroMatch paper for more detail \cite{lou2020neural}.\looseness=-1
When the model is trained, we pre-compute and store embeddings of all graphs in the database. The inference process simply iterates through all pairs of query and target nodes, and utilizes the (trained) comparator to make local subgraph decisions. The aggregated decision is then made by checking whether each query neighborhood finds a match. This process has linear complexity in terms of both query and target number of nodes, thus facilitates efficient retrieval at the front-end interface. \looseness=-1


\subsection{Node Alignment via NeuroAlign}
\label{sec:neuroalign}

NeuroMatch determines whether the query is a subgraph of the target graph. When a matching target graph is retrieved and visualized, it is still difficult for the user to extract insights when the target graph is large and the topology is complex. In this case, showing the corresponding nodes can provide intuitive and explainable visual cues. We propose \algorithmname, to obtain improved node alignment performance. We formulate the prediction problem as a classification task, where query nodes are examples and the target nodes correspond to labels. This architectural change is crucial to enable more accurate alignment by accounting for much larger areas on both graphs. However, for different target graphs, the number of classes (i.e.\ target nodes) varies. This creates a challenge for predictive models. We resolve it by employing a flexible, cross-graph attention mechanism.\looseness=-1


As shown in \autoref{fig:neuroalign}, \algorithmname\ directly takes the node embeddings obtained from GNN on the entire graphs $G_Q$ and $G_T$. These embeddings are denoted as $\{h_q,\forall q\in G_Q\}$ and $\{h_t,\forall t\in G_T\}$. We then compute the similarity between each query embedding and every target embeddings through an attention network. This process can be considered as creating an attention matrix $\mathbf{A} \in\mathbb{R}^{\|V_Q\|\times\|V_T\|}$, where the element $\mathbf{A}_{q,t}$ contains the attention from node $q$ to $t$. We then directly transform the similarity matrix to a probability matrix $\mathbf{P}\in\mathbb{R}^{\|V_Q\|\times\|V_T\|}$ using row-wise softmax and use them in the cross-entropy loss. Formally,\looseness=-1
  
\begin{equation}
\label{eq:neuroalign}
\begin{gathered}
\mathbf{A}_{q,t}=\psi(h_q\mathbin\Vert h_t) \\
\mathbf{p}_q=\text{softmax}(\mathbf{a}_q) \\
L(G_Q,G_T)=-\sum_{q\in G_Q} \mathbf{y}_q \log(\mathbf{p}_q) 
\end{gathered}
\end{equation}
where $\psi$ denotes the attention network, $\mathbf{a}_q$ is the $q$-th row of $\mathbf{A}$, and $\mathbf{y}_q$ is the one-hot ground-truth label for node $q$, indicating which node in $G_T$ is the corresponding node of $q$. The prediction $\mathbf{p}_q$ contains the probabilities of matching query node $q$ to every target node. We implement the attention network as a multi-layer perceptron, which takes a pair of embeddings produced by the GNN, concatenate them and return a similarity score between a node $q$ in the query graph and a node $t$ in the target graph. In case $G_T$ is too large, the computation of $\mathbf{A}_{q,t}$ could consume too much memory, and needs to be constrained to a subgraph at $t$. In practice, we specify a maximum size that covers most target graphs in the database. \looseness=-1

Similar to NeuroMatch, when the model is trained, we can pre-compute all graph embeddings generated by NeuroAlign to make the retrieval process efficient. In addition, \algorithmname\ works subsequently to NeuroMatch and only activates when a subgraph relationship is predicted, thus creating minimal computational overhead for visualization and interaction.\looseness=-1

\subsection{{Algorithm Training}}
\label{sec:training}

The training of NeuroMatch and NeuroAlign are conducted separately. Training NeuroMatch (and its backbone GraphSAGE GNN) involves sampling large amounts of mini-batches containing both positive and negative pairs. A positive pair consists of two neighborhood graphs $g_q$ and $g_t$ that satisfy the subgraph relationship, while a negative pair consists of neighborhood graphs where the relationship is violated. To sample a positive pair, we first randomly sample a $k-$hop neighborhood as $g_t$, and then sample a subgraph within $g_t$ as the query neighborhood $g_q$. To sample negative pairs, we start with the obtained target neighborhood $g_t$ above, and sample a smaller neighborhood from a different graph as $g_q$ (query neighborhood). Note that $g_q$ needs to be verified with exact matching protocol \cite{cordella2004sub} to ensure $g_q\nsubseteq g_t$. In practice, we find that \textit{hard} negatives are necessary to achieve high precision, which are obtained by perturbing the above positive pair ($g_q\subseteq g_t$) such that the subgraph relationship no longer exists. We perturb the positive pair by randomly adding edges to $g_q$ and verify the success with exact matching \cite{cordella2004sub}. As can be seen, negative sampling extensively invokes exact matching algorithm, which is slow to compute. To keep the training tractable, we set small neighborhood hop $k=3$ and also limit the number of nodes to sample from the neighborhood to $30$. \looseness=-1

Training \algorithmname\ (and its backbone GraphSAGE GNN) is much simpler. It involves sampling only positive pairs, since its objective is to improve node alignment when the subgraph decision has already been made that $G_Q\subseteq G_T$. Therefore, the sampling involves extracting random queries from the graphs in the database. For each target graph $G_T$ in the database, we randomly sample a subgraph within it as $G_Q$. The ground-truth injection mapping is acquired directly in the sampling process, and it is converted to $\mathbf{y}_q$ to indicate which node in $G_T$ is the corresponding node of $q$. \algorithmname\ can be trained efficiently through this simple sampling process and without invoking the expensive exact matching algorithm.\looseness=-1

\subsection{Greedy Assignment for Inference}
\label{sec:assignment}
{During inference of node alignment, different nodes in the query graph could be mapped to the same node on the target graph. This is likely to occur among nodes with highly similar topological and attribute features.} The prediction conflict can be resolved with a task assignment algorithm. Instead of resorting to the combinatorial Hungarian algorithm \cite{munkres1957algorithms}, we further develop a simple greedy assignment approach. Specifically, given the predicted probability matrix $\mathbf{P}$, we iterate the probabilities in descending order and record the corresponding matching pair only when both the query and target nodes have not been assigned. The iteration stops when all query nodes have been assigned. This simple process resolves conflicting assignment to the same target node and improves the overall node alignment performance (experimental results in Section \ref{sec:results_acc}).\looseness=-1

\begin{figure*}[ht]
    \centering
    \vspace{-0.15in}
    \includegraphics[width=0.95\linewidth]{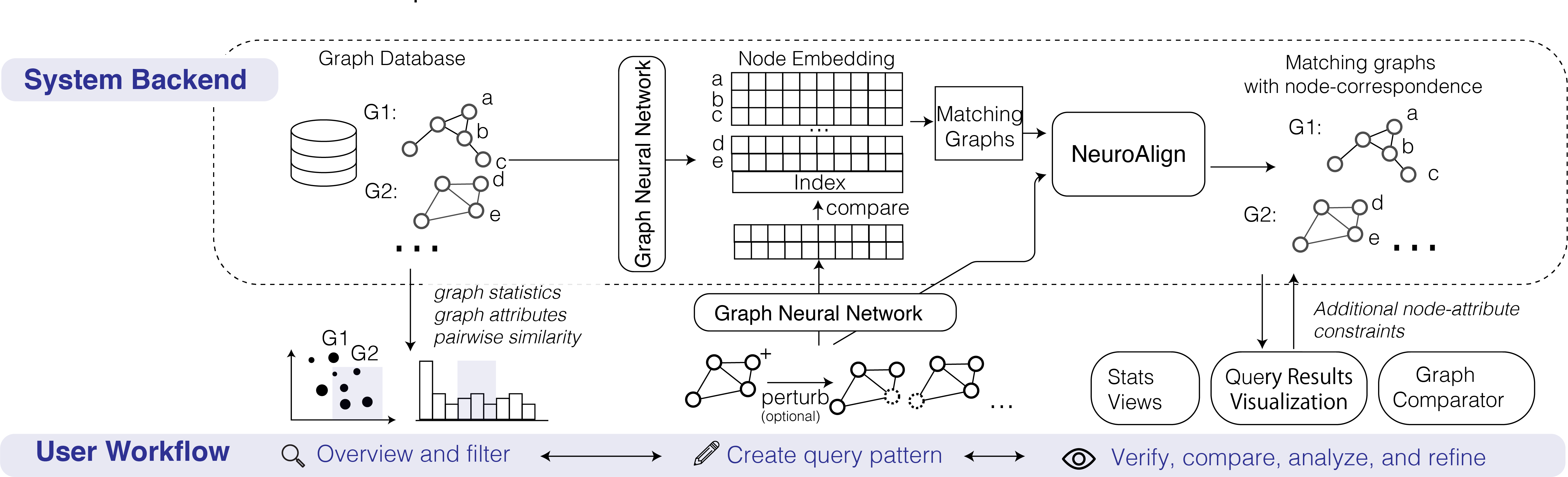}
    \vspace{-0.12in}
    \caption{System architecture of \systemname. The back-end precomputes and stores the graph representations to support efficient matching graph retrieval through the NeuroMatch algorithm. After the matching graphs are obtained, we use \algorithmname~to obtain accurate node-to-node 
    correspondence to be displayed in the visualization for the user to verify the results. Users can start from an overview of all the graphs in the database and select one to construct example-based query pattern. The query pattern can be slightly perturbed to retrieve approximate matching results from the database. After the results are returned, the user can use a variety of views to explore the returned results.\looseness=-1}
    \vspace{-0.20in}
    \label{figure:system_architecture}
\end{figure*}

\subsection{Approximate Query Matching}

In addition to the retrieval results obtained from the query graph, we provide the option to perform approximate query matching. This method perturbs the query graph slightly, in order to obtain similar, but different matching graphs. Specifically, denote the set of obtained matches from the original query graph $G_Q$ as $R$. We remove one node from $G_Q$ and its associated edges to obtain the perturbed query $G'_Q$. {Then we conduct the search with NeuroMatch on $G'_Q$ and add the novel matches $R$. We continue the iteration by removing a node from the perturbed query, until either a prespecified maximum number of steps is reached or $G'_Q$ becomes disconnected. To lower the chance of getting a disconnected graph, each time we remove the node with the lowest degree in $G'_Q$.}\looseness=-1

\section{{Visualization and Interaction}}
{In this section, we first evaluate the design goals of GraphQ (Section \ref{section:design_requirements}). We then describe the GraphQ system with details on its visualization and interaction components (Section \ref{subsection:sys_components}), and technical implementation (Section \ref{subsection:sys_implementation}).}\looseness=-1

\subsection{Design Goals}
\label{section:design_requirements}

\systemname's principle design goal is to provide a generic solution for interactive graph pattern search on a graph database based on user-specified examples. The basic requirement is that the user needs to be able to interactively select and refine graph patterns and analyze the retrieved results. In the meanwhile, the system should display the matching instances as well as explaining the results by highlighting the node correspondences. \looseness=-1

We further enrich and refine the design goals by collecting requirements for domain-specific usage scenarios. We analyzed two example usage scenarios including workflow graph pattern analysis and semantic scene graph analysis in image understanding. For the first usage scenario (details in Section~\ref{subsection:workflow_analysis}) we worked closely with the domain experts who provided the workflow graph data and who are also the end-user of the system. In the second usage scenario, we reference the relevant literature in computer vision on semantic scene graphs. Semantic scene graph is a commonly used graph structure that describes not only the objects in an image but also their relations \cite{johnson2015image}. They are frequently used to retrieve images with the same semantics. By analyzing the commonalities of the two usage scenarios we identified the following user analysis tasks to support in \systemname: \looseness=-1
\vspace{-0.08in}
\begin{enumerate}[label=\textbf{\textit{T\arabic*}}, leftmargin=*]
    \setlength\itemsep{-0.2em}
    \item \label{req:t1} \textbf{Browse/search the graph database}. To start the query process, the user needs to be able to select from hundreds to thousands of graphs.  Therefore, the system should provide graph search and filtering functionalities based on the category, the name, or graph statistics such as the number of nodes/links. Besides that, {a visualization showing an overview of all graphs in the database will be useful to help locate interesting graphs or clusters.} \looseness=-1
    \item \textbf{Interactively construct the query pattern} by selecting on a graph visualization. To minimize user effort, the system should support both bulk selection mechanisms such as brushing the graph regions as well as query refinement methods to add/delete individual nodes/edges from the pattern. \looseness=-1
    \item \textbf{Interpret and validate the matched graphs} via highlighted similarities and differences. To help users interpret the matching results, the node correspondences, as well as differences in the query results, should be highlighted. Furthermore, since the subgraph matching and node correspondence calculation algorithms are not 100\% accurate, the results need to be presented in a meaningful way for easy verification. \looseness=-1
    \item \textbf{Explore the distribution of the matching instances}. {After the matched graphs are returned, the system should indicate how frequently the query pattern occurs in the entire database, and provide the distribution of the pattern among different categories of graphs in the database.}\looseness=-1
    \item \textbf{Refine query results}. A flexible query system should further support query refinement mechanism where the users can apply their domain knowledge to filter the results with additional constraints, such as matching additional node attributes or limiting the results to a certain category of graphs. \looseness=-1
\end{enumerate}

\subsection{GraphQ System}
\label{section:system}

We design \systemname~to support the user analysis tasks (\textbf{T1-5}) described in Section~\ref{section:design_requirements} with the architecture and user workflow featured in ~\autoref{figure:system_architecture}. The user can start with an overview of the graph database (\textbf{T1}), brush, and select a graph to create example-based query patterns (\textbf{T2}). The query pattern (along with optionally perturbed query pattern for approximate query matching) will be sent to the back-end, its node representations will be computed and compared with the precomputed node embeddings to obtain a set of matching graphs containing the query pattern. The matching results along with the query pattern will go through \algorithmname~to compute one-to-one node correspondence. The query results will be displayed in the front-end with multiple levels-of-detail (\textbf{T3}) and can be refined further by adding node-attribute constraints interactively in the query panel (\textbf{T5}). {The distribution of the matching graphs will be highlighted interactively in the database overview panel (\textbf{T4}).}\looseness=-1

\subsubsection{Components}
\label{subsection:sys_components}
The user interface of  \systemname~is composed of four main components:\looseness=-1

\textit{\textbf{Overview and filters}}. In the overview panel (\autoref{fig:teaser}(3)) the system displays the distribution of key graph statistics such as the number of the nodes/edges as well as domain-specific attributes such as the category of the graph. Both univariate distributions and bivariate distributions can be displayed as histograms or scatterplots. Users can brush the charts and select a subset of graphs to create example-based query patterns. 
To provide an overview of the graph structural information and help users navigate and select a graph to start the query {(\textbf{T1})}, we further precompute the graph editing distance \cite{gao2010surveyged} which roughly captures the structural similarities between all pairs of graphs. A 2-D projection coordinates of the graph can then be precomputed using t-SNE \cite{van2008tsne} based on the distance matrix and stored as additional graph attributes (\autoref{fig:teaser}(a)).\looseness=-1

After the query result is obtained, the charts will be updated to provide a contextual view of how the subgraph pattern occurs in the database. For example, the user can observe whether the pattern occurrence concentrate on a small subset of graph categories or it is a generic pattern that appears in many different categories (\textbf{T4}) (\autoref{fig:teaser}(d)).\looseness=-1

Furthermore, the overview panel is a customizable module that can be configured through a json file specifying the attributes to be displayed and the chart to display it. Users can also interactively fold each chart and hide it in the display, such that space can be used for keeping important attribute information on the screen. The system also displays a popup window to show detailed information for selected charts.\looseness=-1

\textit{\textbf{Graph query panel}}. In the graph query panel (~\autoref{fig:teaser}(1)), the user can interactively select from a graph instance to construct the query pattern. The color of the nodes encodes the key node attribute to be matched in the subgraph pattern query. The system currently supports categorical node attributes. This can be extended to numerical attributes by quantizing the values. Additional node attributes are displayed in attachment to the nodes or in tooltips. {As discussed in \autoref{section:design_requirements}, we need to support fast, interactive query construction (\textbf{T2}).} In this panel, the user can quickly select a group of nodes and the subgraph they induce by brushing a rectangular area on the visualization. They can also construct the pattern in a more precise manner by clicking the \textbf{+} and \textbf{-} button on the top right corner of each node. A minimap on the bottom right of the panel allows the user to easily navigate and explore graphs of larger size. The layout of the graph is computed with existing layout algorithms, such as the algorithm described in \cite{gansner1993technique} for directed graphs. When the nodes have inherent spatial locations, they are used directly for display.\looseness=-1

\textit{\textbf{Query results}}. After the sub-graph pattern matching results are returned, the query results panel will be updated to display all the matching graphs as a small multiples display (\autoref{fig:teaser}(2.1) and (2.2)). Since the number of returned results could be large, the system supports sorting the returned graphs with graph attribute values such as the number of nodes (\autoref{fig:teaser}(f)). {To support \textbf{T3}, the matching nodes are highlighted based on the results returned by the node alignment module.} The graphs can be displayed either in a node-link diagram with the same layout as the graph in the query panel (\autoref{fig:teaser}(2.2)) or in a thumbnail visualization designed to display the graph in a more compact manner (\autoref{fig:teaser}(2.1)). In particular, we use topological sorting of the nodes for directed acyclic graphs to order the nodes, layout them vertically, and route the links on the right to obtain a compact view (\autoref{fig:teaser}(2.1)).\looseness=-1

\textit{\textbf{Comparison view}}. {To support \textbf{T3} and \textbf{T5}, we further visualize the query and selected matching graphs side-by-side in a popup window.} The user can click on the zoom-in button on each small multiple to bring out the comparison view (\autoref{fig:teaser}(5)) and review each matching graph in detail. The matched nodes are highlighted for verification.\looseness=-1

\subsubsection{Implementation}
\label{subsection:sys_implementation}

\systemname's implementation uses a typical client-server architecture. The frontend UI framework is implemented in Javascript with React\cite{react} and AntD UI\cite{antd} libraries. The visualizations are drawn using D3.js\cite{d3} on svg within the React framework. We use dagre \cite{dagre} to compute directed graph layout in the front-end. The backend server is implemented in Python with Flask \cite{grinberg2018flask}. The graph data are stored as json documents in the file system and modeled with NetworkX \cite{hagberg2008exploring}. We use PyTorch \cite{NEURIPS2019_9015} for graph representation learning for both subgraph matching and node correspondence learning. More specifically, we use PyTorch Geometric \cite{torch_geometric} and DeepSNAP \cite{deepsnap} to batch graph data (including their topological structures and node features) for training and inference. \looseness=-1

\section{Evaluation}

{Our evaluation of the proposed system consists of two example usage scenarios (Section \ref{subsection:workflow_analysis} and \ref{subsection:scene_graph}), quantitative experiments on various datasets (Section \ref{subsection:experiment_results}), and interview with domain experts on both usage scenarios (Section \ref{subsection:expert_interview}).}\looseness=-1

\subsection{Example Usage Scenario: Program Workflow Analysis}
\label{subsection:workflow_analysis}

\begin{figure*}[h]
    \centering
    \vspace{-0.15in}
    \includegraphics[width=0.9\linewidth]{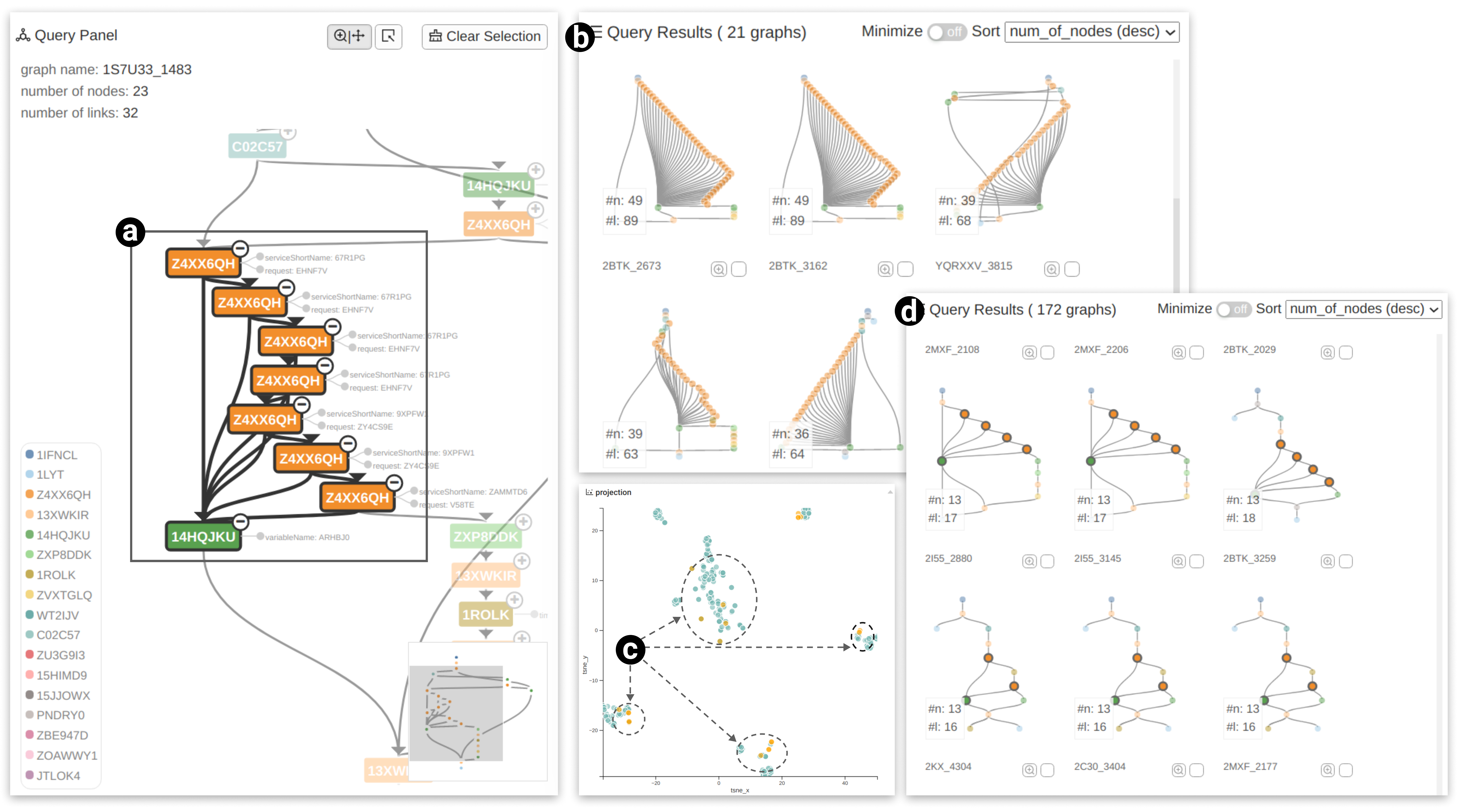}
    \vspace{-0.15in}
    \caption{The user selects a fan-like pattern (a). Exact subgraph matching returns 21 results (b). After enabling approximate search (\autoref{fig:teaser}(4)), the back-end returns 172 graphs (d) containing fan-like patterns, although some of them are simpler than the query. The query results indicate that such structure can be reused as a template to reduce the manual effort for future workflow creation.\looseness=-1}
    \vspace{-0.10in}
    \label{fig:case_study_1}
\end{figure*}

\begin{figure}[h]
    \centering
    \vspace{-0.10in}
    \includegraphics[width=\linewidth]{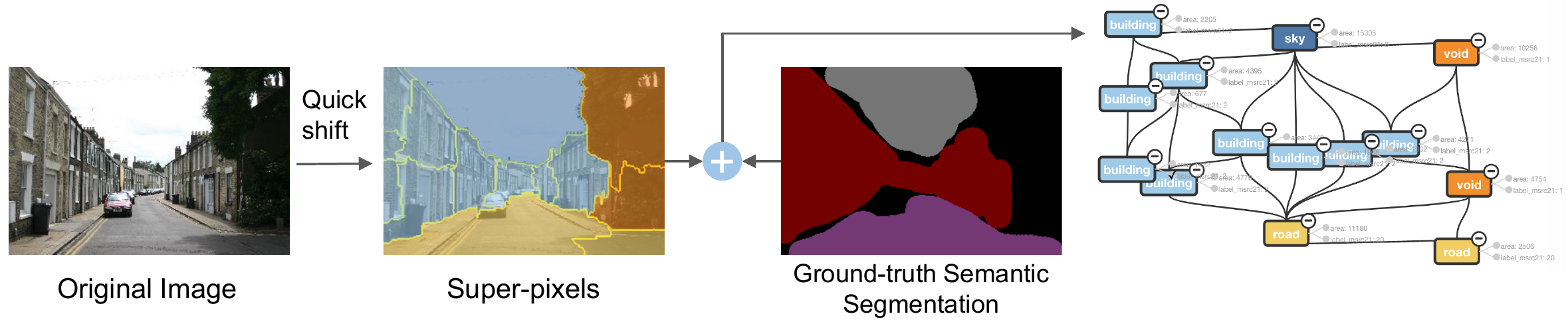}
    \vspace{-0.2in}
    \caption{
    To obtain a semantic scene graph from an image in the MSRC-21 dataset, we use the Quickshift \cite{vedaldi2008quick} algorithm which segments the image into partitions, i.e. super-pixels; then we derive each semantic label as the most frequent ground-truth label of all pixels inside the corresponding super-pixel. Each super-pixel is mapped to a graph node with the semantic attribute. \looseness=-1}
    \vspace{-0.2in}
    \label{figure:scene_graph_extraction}
\end{figure}

In the first usage scenario, we apply \systemname~to analyze a collection of graphs describing the workflows in a vehicle diagnostics software program. The software program uses prescripted workflow graphs to check the functionalities of the system and locate the problem in the vehicles. The workflows are modeled as directed graphs where each node represents an individual procedure in the workflow and the link represents their sequential orders. {We convert the graphs to undirected graphs as input for the query algorithms.} In total, there are $\sim$20 different types of procedures in the workflow, and we use node colors in the system to distinguish them (\autoref{fig:teaser}) (all the names of the nodes are anonymized). In both NeuroMatch and \algorithmname, the type of the procedures is considered as a node attribute. \looseness=-1

The workflows are manually created and it is a time-consuming process. The goal of analyzing workflow graphs is to identify subroutines in the workflow that are reused frequently and therefore can be used as templates, or submodules in the future to facilitate the workflow editing process or to simplify the workflow descriptions. However, identifying such frequent subroutines cannot be easily automated -- substantial domain knowledge in automotive hardware and software system is needed to curate meaningful patterns, therefore a human-in-the-loop approach is well-suited. \looseness=-1

{Through an initial data exploration together with the domain experts, we found that pairwise comparison of workflows using graph editing distance \cite{gao2010surveyged} can provide an overview of the graph similarities in the dataset. This overview can help the user to select interesting workflows as the starting point for exploration. Our system integrates a t-SNE projection \cite{van2008tsne} of all the graphs based on the graph editing distance matrix which reveals several clusters (\autoref{fig:teaser}(a)). The user can use the brushing function to select one cluster and the selected graphs will be updated in the table (\autoref{fig:teaser}(b)). The user could then select any graph from the table to be displayed in the query editor (\autoref{fig:teaser}(1)) to create example-based queries.} In \autoref{fig:teaser}(c), a subroutine with a branching structure is selected by brushing on the visualization. The user can invoke the context menu and search for the query pattern in the graph database. With approximate matching disabled (\autoref{fig:teaser}(4)), the system returns 45 matched graphs in the database. In the graph types histogram, we can see that most of the matched graphs belong to two types (\autoref{fig:teaser}(d)). For an overview of the matching results (\autoref{fig:teaser}(2.1)), the user could toggle minimize in the query results display (\autoref{fig:teaser}(f)) and highlight the node matches returned by \algorithmname\ (\autoref{fig:teaser}(e)). The result shows that indeed most of the graphs returned contain the nodes in the query pattern, indicating that the algorithm is returning reliable results. To further view the details, the user turns off the minimize toggle, and the graphs are displayed in a similar layout as in the query panel and the user can review more details about each graph including the graph name, number of nodes, and links, etc (\autoref{fig:teaser}(2.2)). {To facilitate the inspection of more detail about the returned matches and aligned nodes, we design the side-by-side display of the query graph and returned matching graph (\autoref{fig:teaser}(5)). The display is activated as a popup window when the user clicks on the zoom button (\autoref{fig:teaser}(g)).} Users can also add additional node attribute constraints by clicking on the corresponding node attribute (\autoref{fig:teaser}(h)) to be matched in the query results. In this example there is no workflow satisfying the specified attribute constraint. After verifying the results the user can save the query pattern in a json file to be reused when manually creating workflows in the future.\looseness=-1


\autoref{fig:case_study_1} shows the query results for a fan-like structure selected from a graph (\autoref{fig:case_study_1}(a)). The system returns 21 matched results with approximate search disabled. Indeed most of the returned graphs contain the fan-like structure (\autoref{fig:case_study_1}(b)), indicating another reusable submodule in the workflow creation process. In the t-SNE plot, the graphs with matching fan-like patterns are highlighted in orange, showing the graphs are scattered in different clusters according to graph editing distance (\autoref{fig:case_study_1}(c)). {This finding indicates our method can uncover meaningful patterns in the sub-regions of the graphs that are missed by graph-level similarities.} To further extend the search to graphs that may contain similar, but not exact the same patterns, the user toggles the button to enable approximate search (\autoref{fig:teaser}(4)), the returned result contains much more graphs  (172 graphs) than in exact matching (\autoref{fig:case_study_1}(d)). The user sorts the results based on the number of nodes and found that the graphs with approximate matches contain a simpler fan-like structure with fewer nodes. Based on the analysis the user concludes that the fan-like pattern can be used as a template in the future. \looseness=-1

\begin{figure*}[th]
    \centering
    \vspace{-0.10in}
    \includegraphics[width=0.95\linewidth]{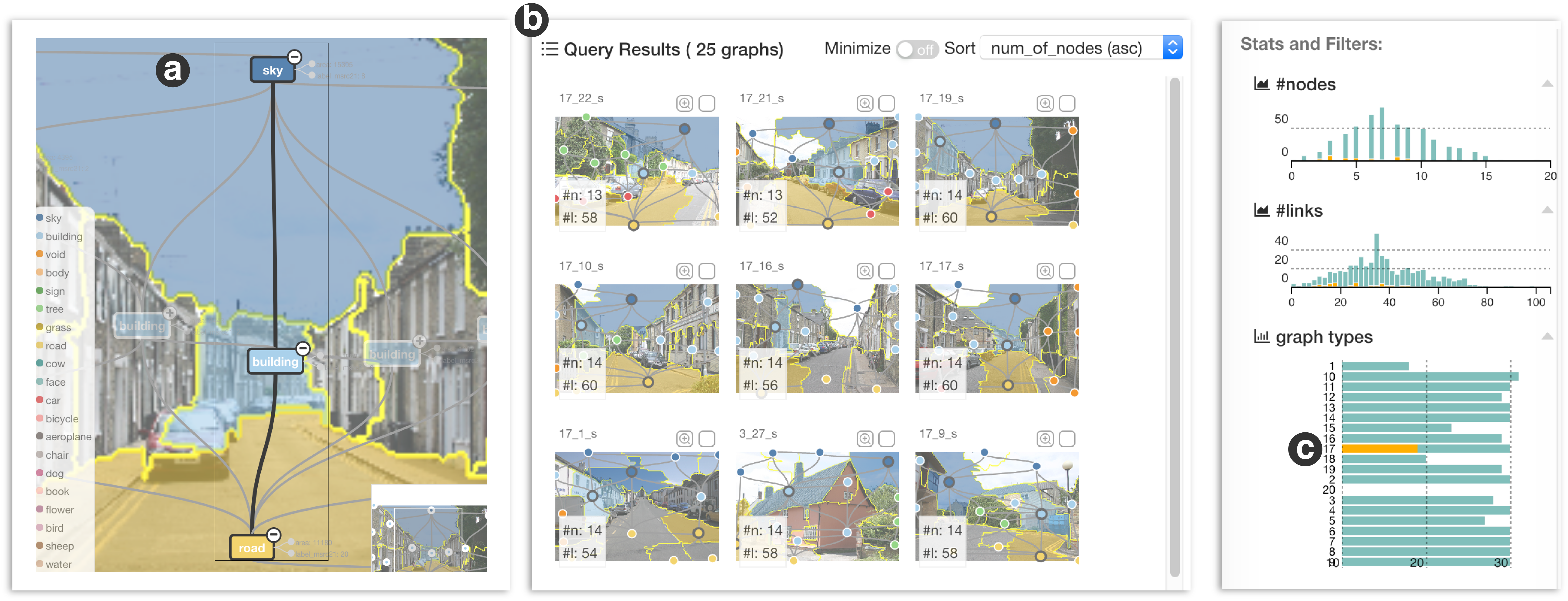}
    \vspace{-0.10in}
    \caption{Case study 2, searching by brushing a subregion (a chain of sky, building, and road nodes) on the (MSRC-21) scene graph and find the matching results (b), most of which contain the same chain of such three nodes as in (a). The three nodes' relationship resembles a typical street view image. \looseness=-1}
    \vspace{-0.10in}
    \label{figure:case_study_2}
\end{figure*}

\subsection{Example Usage Scenario: Scene Graph Search}
\label{subsection:scene_graph}

In the second usage scenario, we apply \systemname~to semantic scene graph search in computer vision applications to find images with similar objects and relationships that resemble our query subgraph structure. {It can be useful for many computer vision tasks such as image retrieval \cite{schroeder2020structured,yoon2020image}, visual question answering, relationship modeling, and image generation.} We follow the procedures described in \cite{propagationkernels} to extract a semantic scene graph from each image. Each node in the graph represents a super-pixel extracted from the image using a segmentation algorithm and the links between nodes encode the adjacency information between those super-pixels. Each node is annotated with a semantic label, as one of its attributes and the whole extracted graph from an image is an undirected, planar graph \cite{planargraph}. In this study, we use a public image segmentation dataset (MSRC-21 \cite{msrc21}) to illustrate this approach. Each image contains ground-truth labels such as \textit{tree}, \textit{grass}, \textit{wall} and unlabeled \textit{void}, etc. We illustrate the process to extract the scene graph from each image in \autoref{figure:scene_graph_extraction}. \looseness=-1

To perform scene graph search, the user starts with the overview of all graphs in the database. The user picks a graph to work on and brushes a subgraph, for example, three connected nodes (\autoref{figure:case_study_2}(a)) including sky, building and road. This subgraph structure could indicate a typical city environment (with buildings and road). The backend, with approximate search disabled, returns matched result of 25 graphs and most of them contain the same subgraph: street view: interconnected super-pixels of sky, building and road as shown in  (\autoref{figure:case_study_2}(b)). Note in histogram overview (\autoref{figure:case_study_2}(c)), all of these resulted images come from the same row (17th) in MSRC-21 dataset that belongs to the category ``road/building". The user can also sort by different metrics and filter by different node information such as area range, or even super-pixel location, etc. Through these interactions, the user eventually finds interesting images tailored to needs.\looseness=-1


\subsection{Quantitative Evaluation}
\label{subsection:experiment_results}

\begin{table}[t]
\centering
\caption{Subgraph decision performance using NeuroMatch.}
\begin{tabular}{cccc}
\hline
\textbf{Dataset}  & \textbf{Precision} & \textbf{Recall} & \textbf{F1} \\ \hline
\textbf{Workflow} & 87.0               & 89.9            & 88.4        \\ 
\textbf{MSRC-21}  & 83.6               & 91.6            & 87.4        \\ 
{\textbf{COX2}}  & 87.4               & 90.9            & 89.1        \\ 
{\textbf{Enzymes}}  & 81.8               & 73.0            & 77.1        \\ \hline
\end{tabular}
\vspace{-0.20in}
\label{tab:subgraph}
\end{table}

\begin{table*}[t]
\centering
\vspace{-0.10in}
\caption{Node alignment performance. \algorithmname~achieves averaged 25\% improvement on the final accuracy.}
\begin{tabular}{c|ccccc|ccccc}
\hline
\textbf{Method}            & \textbf{Dataset}                   & \textbf{\begin{tabular}[c]{@{}c@{}}top-1 \\ acc.\end{tabular}} & \textbf{\begin{tabular}[c]{@{}c@{}}top-2 \\ acc.\end{tabular}} & \textbf{\begin{tabular}[c]{@{}c@{}}top-3 \\ acc.\end{tabular}} & \textbf{\begin{tabular}[c]{@{}c@{}}acc. w/ \\ assignment\end{tabular}} & \textbf{Dataset}                  & \textbf{\begin{tabular}[c]{@{}c@{}}top-1 \\ acc.\end{tabular}} & \textbf{\begin{tabular}[c]{@{}c@{}}top-2 \\ acc.\end{tabular}} & \textbf{\begin{tabular}[c]{@{}c@{}}top-3 \\ acc.\end{tabular}} & \textbf{\begin{tabular}[c]{@{}c@{}}acc. w/ \\ assignment\end{tabular}} \\ \hline\hline
\textbf{NeuroMatch}        & \multirow{2}{*}{\textbf{Workflow}} & 64.2                                                           & 85.6                                                           & 93.4                                                           & 68.6                                                                   & \multirow{2}{*}{{\textbf{COX2}}}    & 42.2                                                           & 56.5                                                           & 65.9                                                           & 44.1                                                                   \\
\textbf{NeuroAlign (Ours)} &                                    & \textbf{91.5}                                                  & \textbf{97.7}                                                  & \textbf{98.7}                                                  & \textbf{95.2}                                                          &                                   & \textbf{65.3}                                                  & \textbf{81.6}                                                  & \textbf{92.0}                                                  & \textbf{70.4}                                                          \\ 
\hline
\textbf{NeuroMatch}        & \multirow{2}{*}{\textbf{MSRC-21}}  & 40.9                                                           & 62.7                                                           & 77.0                                                           & 52.6                                                                   & \multirow{2}{*}{{\textbf{Enzymes}}} & 41.7                                                           & 56.6                                                           & 67.4                                                           & 47.5                                                                   \\
\textbf{NeuroAlign (Ours)} &                                    & \textbf{59.6}                                                  & \textbf{84.2}                                                  & \textbf{95.1}                                                  & \textbf{81.3}                                                          &                                   & \textbf{53.6}                                                  & \textbf{75.3}                                                  & \textbf{86.3}                                                  & \textbf{66.7}                                                          \\ 
\hline
\end{tabular}
\vspace{-0.20in}
\label{tab:align}
\end{table*}

We evaluate the performance of the proposed system on {4 graph datasets in various domains}: program workflow dataset (vehicle diagnostic), MSRC-21 (image processing), COX2 (chemistry) and Enzymes (biology). {The workflow dataset contains $\sim$500 individual workflow graphs with the number of nodes ranging from 5 to 150. $\sim$20 different types of nodes correspond to different diagnostic procedures. MSRC-21 \cite{msrc21} contains natural scene images with 21 object semantic labels. After the super-pixel extraction and processing steps as described in Section \ref{subsection:scene_graph} and \autoref{figure:scene_graph_extraction}, the resulting graph dataset includes 544 graphs with 11 to 31 nodes. COX2 \cite{sutherland2003spline,Morris+2020} consists of 467 chemical molecule graphs with the number of nodes ranging from 32 to 56. Enzymes dataset \cite{schomburg2004brenda,Morris+2020} contains 600 graphs of protein tertiary structure with 3 to 96 nodes. The last 3 datasets are public.}\looseness=-1

{We utilize an 8-layer GraphSAGE in training and the hidden dimension for node embeddings is 64. For NeuroAlign, the attention network has two hidden layers of dimensions 256 and 64. We use ReLU activation. The learning rate is fixed at 0.0001 without weight decay and Adam optimizer is utilized.}\looseness=-1

{The training data is generated on the fly by randomly sampling the positive and negative pairs, as described in \autoref{sec:training}. Note that the ground-truth label for a positive pair is obtained automatically during sampling, and for a negative pair is calculated by exact matching algorithm \cite{cordella2004sub}. The batch size is fixed to 128. For validation data, we sample the dataset following the same process, prior to training. For testing data, we sample based on the evaluation tasks as described in the following sections.}\looseness=-1

All experiments are conducted on a single GeForce GTX 1080 Ti GPU. We measure the performance of the system in terms of prediction correctness and runtime efficiency. For all evaluations, the approximate query matching is turned off. The detailed description of the evaluation setup and experimental results are presented below. \looseness=-1

\subsubsection{Prediction Accuracy}
\label{sec:results_acc}

To construct the testing dataset for evaluation of the prediction accuracy, we randomly extract $5$ queries from each graph, and obtain their ground-truth subgraph-isomorphism labels. The evaluation is conducted on the problem of subgraph decision and node alignment separately. For subgraph decision, we measure the precision and recall, commonly used in the information retrieval domain, to measure how well NeuroMatch retrieves the ground-truth matching target graphs from the graph database. \looseness=-1

For node alignment, the objective is to measure how well the algorithm predicts the correct matching nodes on the retrieved target graphs. Since the wrong retrieval does not have ground-truth node alignment, we conduct the evaluation on the set of correctly retrieved target graphs. For this task, we compare our proposed \algorithmname\ with NeuroMatch, which provides node correspondence through the matched anchor nodes. Greedy assignment (Section \ref{sec:assignment}) is applied on both NeuroMatch and \algorithmname\ to improve the inference. The details on utilizing the greedy assignment on NeuroMatch can be found in the appendix. To measure the performance, we calculate the top-$k$ ($k\in\{1,2,3\}$) accuracy along with the accuracy after the greedy assignment on each query, and report the average among all queries. {In case multiple matches exist in the ground truth, we only consider the one closest to algorithm prediction to measure the accuracy.} The identification of multiple subgraph isomorphisms \cite{liu2020neural} is a more challenging research topic and we provide a discussion in Section \ref{sec:discussion}.\looseness=-1

The performance of subgraph decision is shown in Table \ref{tab:subgraph}. The results show that the system is able to retrieve around $90\%$ matching target graphs for both datasets while maintaining high precision. Note that achieving high precision is much more challenging than high recall since a matching target graph is rare as compared to non-matching graphs. The excellent precision and F1 score of the system demonstrate the model's capability to learn embeddings that correctly reflect the subgraph relationship.\looseness=-1

The comparison between NeuroMatch and our proposed algorithm \algorithmname\ on the node alignment task is shown in Table \ref{tab:align}. NeuroMatch performed poorly on this task due to multiple predicted matches for many query nodes. We achieve significant improvement over NeuroMatch (e.g. $27.3\%$ improvement on top-$1$ acc. and $22.2\%$ improvement after assignment for Workflow, $18.7\%$ improvement on top-$1$ acc. and $28.7\%$ improvement after assignment for MSRC-21). We also observe that MSRC-21 is much more challenging than Workflow dataset due to the dense connectivity and a large number of similar adjacency nodes. Interestingly, although \algorithmname\ makes many wrong decisions from the top-$1$ predictions, its top-$3$ predictions contain most labels. As a result, the simple assignment approach successfully resolves many predicted conflicts and significantly improves the accuracy. Contrarily the assignment does not make much improvement for NeuroMatch predictions. In addition, we experimented with the optimal Hungarian assignment algorithm and observe that, as compared to our greedy approach, the improvement is negligible for \algorithmname, but higher for NeuroMatch (e.g. achieves $73.1\%$ acc. on Workflow and $55.4\%$ acc. on MSRC-21) due to more conflicting predictions.\looseness=-1

\begin{figure}[t]
    \centering
    \includegraphics[width=0.9\linewidth]{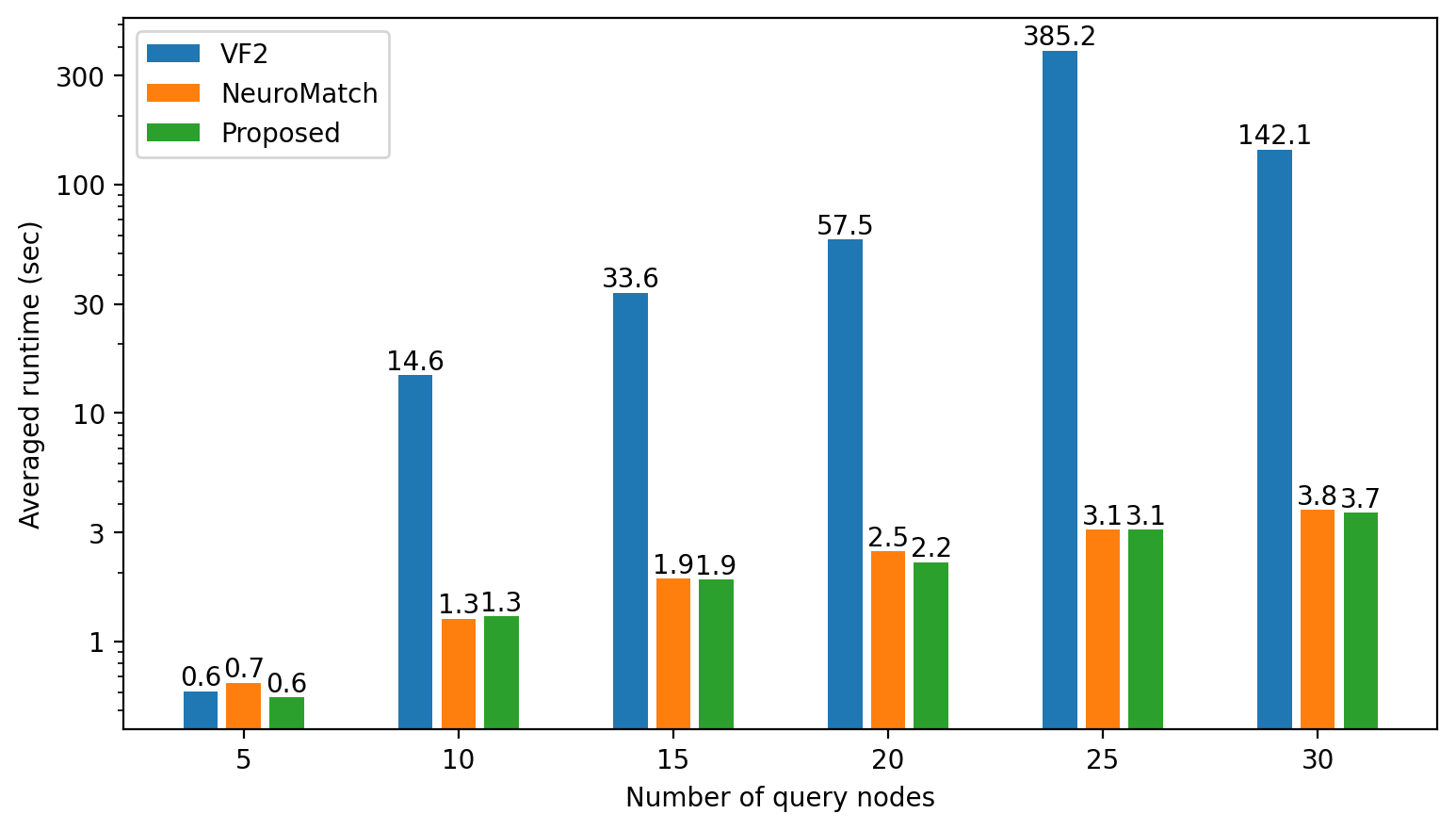}
    \vspace{-0.10in}
    \caption{Runtime comparison with VF2\cite{cordella2004sub} and NeuroMatch\cite{lou2020neural} on the Workflow dataset. Runtime in seconds is shown on the $y$-axis as logarithm scale and the exact number is above the bar. Compared to VF2, our system provides 10$\times$--100$\times$ speedup starting from 10 query nodes and therefore enables interactive query. Our proposed \algorithmname\ component adds little to none computational overhead as compared to NeuroMatch, while providing much more accurate node-alignment results.\looseness=-1}
    \vspace{-0.15in}
    \label{fig:speed}
\end{figure}

\subsubsection{Runtime Efficiency}
\label{sec:results_speed}
{Next, we measure the runtime efficiency in comparison with the VF2 baseline \cite{cordella2004sub} to evaluate the speed gain. VF2 is the state-of-the-art exact matching algorithm based on backtracking procedure. Although it calculates true subgraph-isomorphism results, the computation is expensive, especially for larger graphs.} In addition, we also compare with a similar system where \algorithmname\ component is removed to evaluate the added computational overhead of \algorithmname. For this evaluation, we consider the number of query nodes ranging from $5$ to $30$ with an increment of $5$ on the Workflow dataset, and randomly extract $2000$ corresponding queries for each number. We measure the averaged runtime in seconds for the matching with the entire database. The results are visualized in \autoref{fig:speed}. We observe that the runtime of VF2 increases exponentially with the increase in query nodes and reaches close to $6$ minutes with just $25$ query nodes. With further increased query nodes they become larger than many target graphs and cannot be matched, thus creating a runtime drop at node size $30$. In contrast, our runtime increases linearly with query node size. Compared to NeuroMatch, the added \algorithmname\ component induces little to none computational overhead. Surprisingly it is slightly faster than NeuroMatch in some cases. We conjecture this is due to the easier assignment task generated by \algorithmname\ (i.e.\ fewer conflicts), such that the greedy algorithm can terminate early.\looseness=-1

\subsection{Expert Interview}
\label{subsection:expert_interview}

To evaluate the usability of the system, we conducted a semi-structured interview involving three industrial experts working on program workflow construction and review for the first usage scenario, as well as {three researchers} working in the computer vision domain for the second usage scenario. We introduced the system with a walk-through of the interactive features and visual encodings and then explored the system together through a remote call. We report a brief summary of the findings here as an initial validation of the usability and utility of the system.  \looseness=-1

For the first usage scenario, {the domain experts considered the visual analytic system easy to understand and fits their current usage scenario very well: identifying reusable workflow modules to simplify future workflow creation. They can easily create new patterns and search for matching graphs in the database and validate the results in the visualization interface.} They even proposed new usages such as using the visualization to review newly created workflows. {One of them  commented, ``The abstraction and searching of custom queries open up a lot of opportunities".} In addition, they requested that the returned workflows to be grouped by additional node features for fine-grained analysis. We are currently working with the experts to deploy the system for larger-scale use, and are expecting more feedback after long-term usage. \looseness=-1

For the second usage scenario, {the domain experts appreciated the usefulness of the system by commenting, ``It's great to perform query so fast and see results interactively. It's certainly very powerful for many computer vision problems".} They showed great interest in applying the system for diagnosing computer vision models to answer questions such as: does an object detection model performs worse when the object is placed on the road instead of in a room? {One of them is interested in retrieving images containing similar semantic structure as some failure cases of the model to perform further analysis and model refinement. Another expert is interested in utilizing the tool for computer vision problems with a heavy focus on object relationships, such as image captioning and visual question answering.} For improvement, they mentioned that the graph edge could encode additional information such as the relative positions (up, down, left, right) of the superpixels to retrieve similar images. {In addition, a ranking of the matched images could be provided based on the closeness of visual appearance to the query image.} \looseness=-1
\section{{Discussion, Limitations and Future Work}}
\label{sec:discussion}
We introduced a novel system \systemname\ to perform interactive visual pattern queries on graph databases based on user-created query patterns. To facilitate interactive query, we utilize graph representation learning to resolve the problem of subgraph decision and node alignment. The intuitive and explainable visual cues provided by \algorithmname\ are paired with novel visual and interaction designs to help users navigate the retrieval results and extract insights. 
Due to the complexity of the subgraph matching problem, there are still many open questions we have not addressed yet:\looseness=-1

\textbf{Node alignment for multiple subgraph isomorphism.} Currently, the training and inference of \algorithmname\ focus on a single instance of subgraph isomorphism. However, in practice, the query nodes could be mapped to multiple sets of nodes in the same matching target graph. Counting and enumerating all these instances is a very challenging problem and requires future research. {Besides that, multiple pattern matches in a large graph bring additional challenges for interaction and scalable visual representations.} \looseness=-1

\textbf{Scalability to very large query graphs.} During training of NeuroMatch, we observe that hard negative samples are crucial to achieving high precision rate. However, sampled or perturbed queries need to be verified with exact matching algorithms to ensure the subgraph relationship does not exist. These algorithms are slow to compute especially when the query and target neighborhood graphs become larger and the connectivity becomes denser. A potential approach to alleviate the issue is to assign large weights to these hard negatives and reduce the overall need to invoke these algorithms during training.   \looseness=-1

{\textbf{Handling directed or disconnected query patterns.} Currently, our algorithm works with using undirected, connected graphs as the query pattern. For directed graphs, we converted them into undirected graphs as input for NeuroMatch and NeuroAlign. To account for the direction of connectivity, the backbone GNN model needs to be modified. For example, GraphSAGE can be modified by distinguishing the in-node and out-node neighborhoods during the aggregate-update process and other GNNs specifically designed for directed graphs such as \cite{tong2020directed,shi2019skeleton} can be considered. On the other hand, for disconnected query patterns, a potential workaround is to consider each connected component separately and make an ensemble of the individual predictions. However, the performance still needs to be investigated.}\looseness=-1


In the future, besides addressing the aforementioned limitations, we plan to investigate database index applied on the embeddings of the large graph database to allow even more efficient retrieval at sub-linear time. Furthermore, considering the wide variety of graph-structured data, we plan to extend the current work to more usage scenarios including social network analysis \cite{yanardag2015deep} and $3$-D point clouds \cite{neumann2013graph}. \looseness=-1


\clearpage


\bibliographystyle{abbrv-doi}

\bibliography{references}
\end{document}